\documentclass[letterpaper]{article} 
\usepackage[preprint]{aaai2027}  
\usepackage[hyphens]{url}  
\usepackage{graphicx} 
\usepackage{natbib}  
\usepackage{caption} 
\usepackage{algorithm}
\usepackage{algorithmic}

\usepackage{newfloat}
\usepackage{listings}
\DeclareCaptionStyle{ruled}{labelfont=normalfont,labelsep=colon,strut=off} 
\floatstyle{ruled}
\newfloat{listing}{tb}{lst}{}
\floatname{listing}{Listing}

\usepackage{booktabs}
\usepackage{amsmath}
\usepackage{amsfonts}

\title{HERA: Historical Evidence Routing Adapter for Physical Prediction in Latent World Models}
\author{
  Yuanruyi\textsuperscript{\rm 1,\rm 4},
  Yue Cao\textsuperscript{\rm 2,\rm 3,\rm 4},
  Haojia Gao\textsuperscript{\rm 4,\rm 5},
  Guanqiu Guo\textsuperscript{\rm 2,\rm 4,\rm 5},
  Ziyuezhang\textsuperscript{\rm 2},
  Shangqin Lin\textsuperscript{\rm 4},
  Junbo Tan\textsuperscript{\rm 2},
  Bokui Chen\textsuperscript{\rm 5},
  Zhuo Zou\textsuperscript{\rm 3},
  Xueqian Wang\textsuperscript{\rm 5}
}
\affiliations{
  \textsuperscript{\rm 1}Chongqing University, 
  \textsuperscript{\rm 2}Research Institute of Tsinghua University in Shenzhen, 
  \textsuperscript{\rm 3}Fudan University, 
  \textsuperscript{\rm 4}Everwise-Tech Co., Ltd., 
  \textsuperscript{\rm 5}Tsinghua Shenzhen International Graduate School\\
  Yuan Ruyi@stu.cqu.edu.cn
}

\begin{document}

\maketitle

\begin{abstract}
  Predictive video models have emerged as promising world models by learning latent visual dynamics from large-scale video.
  Yet these models remain challenged by physical events under occlusion, where later predictions may depend on object evidence that is no longer available in the current view. Addressing this challenge requires historical evidence not only to be preserved but also to remain accessible when it becomes relevant to a subsequent prediction. Existing approaches mainly enlarge the temporal context, cache generic video features, or impose explicit object-centric states, thereby improving the capacity or structure of retained history. However, they do not directly address how relevant historical evidence can be selectively retrieved and integrated into a pretrained predictor without interfering with its native latent workspace.
  Accordingly, we introduce HERA (Historical Evidence Routing Adapter), a framework for routing retained historical evidence into a frozen latent predictor, and instantiate it with Register-Routed Patch Memory (RRPM), a lightweight adapter comprising a Structured Memory Bank, Memory Registers, and Workspace Registers.
  On the IntPhys2 Main split, HERA with RRPM improves the pairwise AvgSurprise accuracy of V-JEPA 2-G from 52.57\% to 54.35\%. Subgroup analysis shows particularly strong improvements on fixed-camera continuity, from 46.15\% to 57.69\%, and fixed-camera immutability, from 46.15\% to 63.46\%.
  These results support historical evidence routing as a practical adaptation strategy for physical prediction in latent world models.

\end{abstract}


\section{Introduction}
\label{sec:introduction}

An ideal world model should not only follow the next visible frame, but also
anticipate the consequences of events from observations that are incomplete,
transient, and often occluded.
It must keep commitments when the evidence is no longer on screen.
A ball should continue behind an occluder, a block should keep its color after it disappears, and a later collision should still be constrained by objects hidden earlier. We refer to this capacity as physical memory: the ability to preserve and recover evidence about existence, attributes, motion, and interaction that persists through time.

Predictive video representation learning provides a natural framework for this capability. Masked video autoencoders and JEPA-style objectives learn by anticipating missing or future visual content~\citep{he2022mae,tong2022videomae,feichtenhofer2022masked,assran2023self}. V-JEPA 2 provides a particularly instructive example because it predicts future latent representations rather than pixels, and supports understanding, prediction, and planning in physically grounded settings~\citep{bardes2024vjepa,bardes2025vjepa2}.
Such results show that predictive video models can encode useful physical regularities.
These findings do not, however, establish that the model maintains a persistent representation of physical state.
A predictor may exploit recent motion, local appearance, or scene-level correlations without reliably preserving object identity, attributes, or existence through occlusion.

Common strategies for extending temporal reasoning do not resolve this distinction automatically.
A longer input window retains more observations, but sparse early evidence must compete with an increasingly large set of recent tokens.
A generic feature cache preserves history, but does not determine which part of that history should influence a particular future prediction.
Object-centric memory offers a stronger state abstraction, yet depends on stable object discovery, identity binding, and temporal matching, properties that a frozen video representation is not guaranteed to expose.
The unresolved bottleneck is therefore not only whether historical evidence is retained, but whether the relevant evidence can be routed into prediction when it is needed without disrupting the pretrained predictor's native latent workspace.

In this work, we study this mechanism using IntPhys2~\citep{bordes2025intphys2}. The benchmark pairs possible and impossible videos that share scene content but differ in physical validity.
For a predictive model, an impossible continuation should produce a larger
future-latent prediction error, and therefore higher surprise, than its
matched possible continuation. This setting allows us to examine whether
V-JEPA 2 can recover earlier visual evidence when prediction depends on
trajectory or appearance cues that are no longer directly observable.
To make retained historical evidence broadly accessible when it becomes relevant to prediction, we introduce HERA, a framework for routing historical evidence into a frozen latent predictor. We instantiate HERA with Register-Routed Patch Memory (RRPM), which stores structured patch history, retrieves it through dedicated Memory Registers, and uses a separate set of trainable Workspace Registers to integrate the retrieved evidence through the frozen predictor's native self-attention pathway.

Our main contributions are summarized as follows.
\begin{itemize}
  \item We formulate pairwise AvgSurprise as a behavioral probe of physical
    memory and identify evaluation settings in which prediction depends on
    historical evidence that is no longer directly observable.

  \item We introduce HERA with RRPM, a parameter-efficient adapter that
    separates Memory Registers from Workspace Registers while keeping
    the V-JEPA 2 blocks frozen. With only 3.00M
    trainable parameters, it improves V-JEPA 2-G pairwise AvgSurprise accuracy
    from 52.57\% to 54.35\% on the IntPhys2 Main split.

  \item Comparisons with five alternative memory mechanisms and diagnostic
    register removals provide evidence that selective routing offers benefits
    beyond historical storage alone. Subgroup analysis further shows gains of
    11.54\% and 17.31\% on fixed-camera continuity and
    immutability, respectively.
\end{itemize}
Together, these results support selective evidence routing as a tractable component of physical memory, distinct from persistent object-state maintenance.

\section{Related Work}
\label{sec:related}

\subsection{Predictive Video World Models}
\label{subsec:related-predictive}

Self-supervised video learning extends masked image modeling to video, while JEPA-style objectives predict representations rather than pixels~\citep{bao2021beit,he2022mae,tong2022videomae,feichtenhofer2022masked,assran2023self,bardes2024vjepa}.
V-JEPA and V-JEPA 2 predict future latent representations, providing natural test cases for predictive video world models~\citep{bardes2024vjepa,bardes2025vjepa2}.
Recent studies examine the robustness of latent prediction and the emergence of physical variables in large-scale video encoders~\citep{alrasheed2026latent,joseph2026physics}.

Latent world models have long supported imagined rollouts and decision making in reinforcement learning and robotics~\citep{ha2018worldmodels,hafner2019planet,hafner2020dreamer,hafner2023dreamerv3,schrittwieser2020muzero,hansen2022tdmpc}.
More recent video foundation models scale this premise toward physical simulation, embodied prediction, and interactive environments~\citep{bruce2024genie,agarwal2025cosmos}.
However, prediction quality or task success does not establish whether historical physical evidence remains available after it disappears from view.

Violation-of-expectation benchmarks directly probe this distinction through paired possible and impossible events~\citep{baillargeon1985object,riochet2018intphys,piloto2018probing,bordes2025intphys2}.
We use IntPhys2 to test whether V-JEPA 2's predictive pathway preserves and uses the evidence required for physical memory.

\subsection{Long-Horizon Memory for Video Prediction}
\label{subsec:related-memory}

Long-sequence transformers preserve distant information through recurrence, sparse attention, compression, retrieval, or explicit memory tokens~\citep{dai2019transformerxl,beltagy2020longformer,rae2020compressive,borgeaud2021retro,wu2022memorizing,bulatov2022recurrent}.
Video transformers additionally rely on space-time or factorized attention to manage large spatiotemporal token grids~\citep{bertasius2021space,arnab2021vivit}.
Register tokens offer a compact internal workspace, but retaining more context alone does not determine which historical evidence should affect a future prediction~\citep{darcet2024vitregisters}.

Recent video world models support long-horizon consistency through spatial memory, compressed historical latents, adaptive frame compression, and learned context queries~\citep{wu2025spatialmemory,hong2025relic,oshima2025worldpack,yu2026memlearner}.
These methods mainly target generative rollouts; RRPM instead separates Workspace Registers from Memory Registers to route historical patch evidence into a frozen latent predictor.
Object-centric representations provide explicit state abstractions but require
stable object decomposition and temporal association, which frozen video features do not guarantee
~\citep{locatello2020slot,kipf2022conditional,elsayed2022savi}.

\section{Method}
\label{sec:method}
\begin{figure*}[t]
  \centering
  \includegraphics[width=0.9\textwidth]{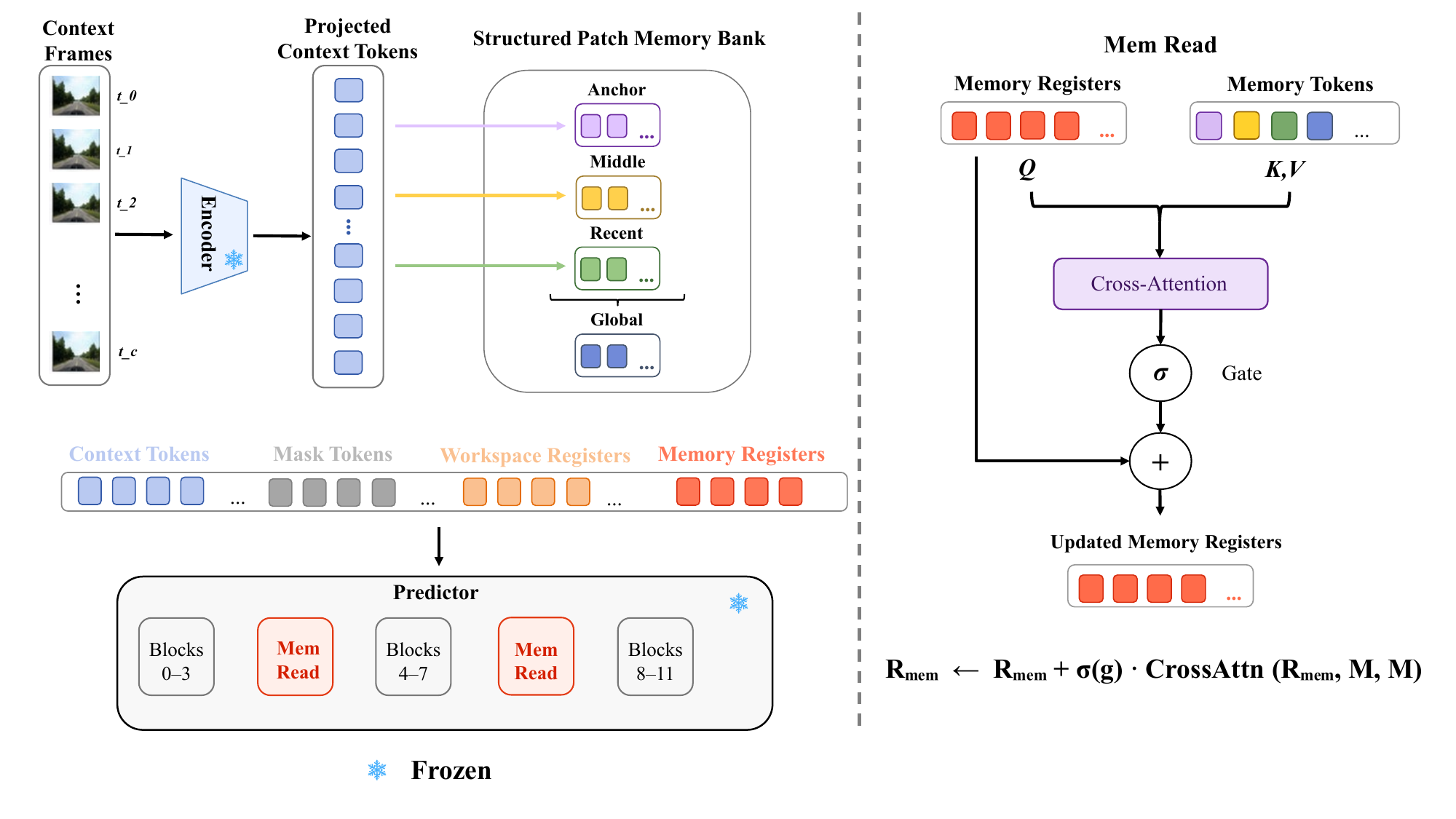}
  \caption{\textbf{Architecture of HERA with RRPM.}
    A frozen V-JEPA 2 context encoder produces tokens that are organized into temporally structured memory.
  Memory Registers retrieve this evidence through gated cross-attention, while Workspace Registers support its integration through the frozen predictor's self-attention pathway.}
  \label{fig:method-overview}
\end{figure*}

Given a matched possible--impossible video pair, we use future-latent prediction surprise to diagnose persistent physical-memory failures.
We then extend a pretrained V-JEPA 2 predictor with HERA without object annotations or changes to its prediction objective.
HERA separates historical retention, selective retrieval, and evidence integration through three RRPM modules: the Structured Memory Bank stores temporally specialized context tokens, the Memory Registers retrieve them through gated cross-attention, and the Workspace Registers integrate retrieved evidence through the predictor's native pathway.
Figure~\ref{fig:method-overview} summarizes the memory construction and routed retrieval process of HERA with RRPM.

\subsection{Mechanism Analysis}
\label{subsec:mechanism-analysis}

\subsubsection{Problem Formulation}
\label{subsec:problem-formulation}

Given a video $v$, V-JEPA 2 observes a context prefix and predicts latent representations of future masked tubelets.
Let $E_{\phi}$ denote the frozen context encoder, $E_{\bar{\phi}}$ the frozen target encoder, and $G_{\psi}$ the frozen predictor.
For a sliding window $w$ with context frames $v_{w,1:c}$ and future target frames $v_{w,c+1:c+h}$, the context encoder produces patch tokens
\begin{equation}
  X_{w,c} = E_{\phi}(v_{w,1:c}) \in \mathbb{R}^{T_c \times P \times D_e},
\end{equation}
where $T_c$ is the number of context tubelets, $P$ is the number of spatial patches per tubelet, and $D_e$ is the encoder dimension.
Let $\Pi:\mathbb{R}^{D_e}\rightarrow\mathbb{R}^{D}$ denote the predictor input
projection, with $\Pi$ equal to the identity map when $D_e=D$, and write
$X_{p,w,c}=\Pi(X_{w,c})$.
The frozen target encoder is applied to the same window and the future mask selects normalized target latents $\bar{Z}_{w,c}$.
The predictor receives $X_{p,w,c}$ and future mask tokens
$Y_{\mathrm{mask}}$, and outputs predicted target latents
$\hat{Z}_{w,c}=G_{\psi}(X_{p,w,c},Y_{\mathrm{mask}})$.

The physical-memory question is whether the model preserves information about objects that are no longer visible but still physically relevant.
We use matched possible/impossible video pairs from IntPhys2~\citep{bordes2025intphys2}.
For scene $s$ and pair index $k$, let $v_{s,k}^{\mathrm{pos}}$ be the physically possible video and $v_{s,k}^{\mathrm{imp}}$ be the physically impossible video.
The two videos share most scene-level factors.
A prediction-based physical model should assign a larger future surprise to $v_{s,k}^{\mathrm{imp}}$.

This formulation separates diagnosis from augmentation.
The diagnostic output is a pairwise surprise margin that tests whether V-JEPA 2 is more surprised by the impossible continuation.
The augmented model changes only the prediction pathway, while the evaluation protocol remains unchanged.
The memory module is trained with the same future-latent objective as V-JEPA 2.
Its physical value is judged only by whether it changes pairwise violation sensitivity.

The official prediction-model scorer first reduces each sliding window to a per-window future-latent prediction loss.
Let $\hat{z}_{v,w,c,t,d}$ denote feature dimension $d$ of predicted target token $t$, and let $\bar{z}_{v,w,c,t,d}$ denote the corresponding normalized target feature.
The per-window surprise is the mean L1 error over target tokens and feature dimensions,
\begin{equation}
  \ell_{v,w,c} =
  \frac{1}{|\mathcal{T}_{v,w,c}|D_z}
  \sum_{t \in \mathcal{T}_{v,w,c}}\sum_{d=1}^{D_z}
  \left|\hat{z}_{v,w,c,t,d} - \bar{z}_{v,w,c,t,d}\right| ,
\end{equation}
where $\mathcal{T}_{v,w,c}$ is the set of predicted future target tokens and $D_z$ is the target latent dimension.
The official video-level score for a fixed context length is therefore the mean over valid, non-padded windows,
\begin{equation}
  \operatorname{AvgSurprise}(v,c) =
  \frac{1}{|\mathcal{W}(v,c)|}
  \sum_{w \in \mathcal{W}(v,c)} \ell_{v,w,c},
\end{equation}
where $\mathcal{W}(v,c)$ is the set of valid windows.
Zero-padded windows are excluded from $\mathcal{W}(v,c)$, and context settings are evaluated independently.
Thus the protocol uses AvgSurprise over all valid sliding windows for a given context length; it does not select a single window and it does not use a max-surprise reduction.

For a matched pair, the pairwise surprise margin is
\begin{equation}
  m_{s,k,c} =
  \operatorname{AvgSurprise}(v_{s,k}^{\mathrm{imp}}, c) -
  \operatorname{AvgSurprise}(v_{s,k}^{\mathrm{pos}}, c).
\end{equation}
The pair is counted as correct only when $m_{s,k,c}>0$; ties are not credited.
The context-wise pairwise accuracy is
\begin{equation}
  \operatorname{Acc}(c) =
  \frac{1}{|\mathcal{P}_c|}
  \sum_{(s,k)\in\mathcal{P}_c}
  \mathbf{1}[m_{s,k,c}>0],
\end{equation}
where $\mathcal{P}_c$ is the set of valid pairs for context length $c$.
For a predefined context set $\mathcal{C}$, the best-context summary is
\begin{equation}
  \operatorname{Acc}_{\mathrm{best}}(\mathcal{C}) =
  \max_{c\in\mathcal{C}} \operatorname{Acc}(c).
\end{equation}
For overall accuracy, the context is selected once at the aggregate
method level and is shared by all pairs. For a subgroup $g$, we analogously restrict the valid pair set to
$\mathcal{P}_{c,g}$ and compute
\begin{equation}
  \operatorname{Acc}_{g,\mathrm{best}}(\mathcal{C})
  =
  \max_{c\in\mathcal{C}}
  \operatorname{Acc}_{g}(c).
\end{equation}
A single context length is shared by all pairs within each method--subgroup
evaluation.
This score is a relative physical-consistency measure rather than an absolute prediction loss.
The possible and impossible videos are scored independently and compared only through the final margin.
No physical labels are used by the memory module during training, and no learned classifier is added at evaluation time.
The probe remains tied to the prediction pathway.
A method succeeds only when the impossible continuation becomes harder for the predictive model to explain than the matched possible continuation.

\subsubsection{Diagnostic Target: Persistent Physical Memory}
\label{subsec:diagnostic-target}

At the behavior level, informative failures depend on early evidence that later
becomes hidden. Solidity requires visible balls or blocks to remain active
through occlusion and contact. Immutability requires static properties to stay
bound to object identity, while permanence and continuity require maintaining
existence and motion without sufficient recent evidence.

At the representation level, these behaviors suggest a gap between patch-level evidence and object-level state.
Frozen V-JEPA 2 latents may preserve local visual traces, but those traces are not necessarily organized as persistent object state.
This motivates temporal-forgetting, attribute-retention, and object-identity
probes that separate missing representation information from failures to route
existing evidence.

At the predictor level, the failure cases suggest over-reliance on recent context.
The predictor can use local continuity and recent patches, but early evidence
can be diluted before it affects future latent prediction.
A memory intervention should preserve early evidence, keep the pretrained
V-JEPA 2 pathway mostly intact, and provide a selective route from stored
evidence into future latent prediction.
These constraints make two simple alternatives unattractive.
Appending all historical tokens directly to the predictor increases context length without deciding which evidence should persist.
Replacing patch tokens with object slots would match the physical target more directly, but it depends on reliable object decomposition from frozen latents.
These observations motivate an intervention that preserves useful
historical evidence, retains the pretrained prediction pathway, and
selectively routes stored evidence into future latent predictions.

\subsection{HERA with RRPM}
\label{subsec:patchmemory-overview}

We extend a pretrained V-JEPA 2 predictor without object annotations or changes to the original latent prediction objective.
HERA separates what historical information should be retained, what should be retrieved for the current prediction, and how retrieved evidence should be integrated.
Its RRPM instantiation contains three modules: the Structured Memory Bank, Memory Registers, and Workspace Registers.

For clarity, we omit the window and context indices below and write
$X_p\equiv X_{p,w,c}\in\mathbb{R}^{T_c\times P\times D}$ for the context
tokens after projection into the predictor hidden dimension $D$.
The original V-JEPA 2 predictor consumes $X_p$ together with future mask tokens $Y_{\mathrm{mask}}$ and predicts target latents.
We keep the context encoder, target encoder, and pretrained predictor blocks frozen.
Only the Structured Memory Bank, cross-attention readers and gates, Workspace Registers, and Memory Registers are trainable.
The memory-augmented predictor therefore has the form
\begin{equation}
  \hat{Z} = G_{\psi}^{\mathrm{mem}}
  (X_p, Y_{\mathrm{mask}}, M, R_w, R_{\mathrm{mem}}),
\end{equation}
where $M$ denotes the Structured Memory Bank, $R_w$ denotes the Workspace Registers, and $R_{\mathrm{mem}}$ denotes the Memory Registers.

The Structured Memory Bank stores selected or compressed patch evidence from different temporal regions of the context.
Workspace Registers support the predictor's internal computation, while Memory Registers retrieve explicit memory and carry it through later self-attention layers.
This separation keeps the intervention small while making early evidence available in a stable form before the frozen predictor is asked to use it.

\subsubsection{Structured Memory Bank}
\label{subsec:patchbank}
Physical prediction often depends on evidence distributed across the context,
with recent observations supporting short-term motion and earlier conditions revealing later physical violations.
Retaining all context patches is redundant and expensive,
while object-centric memory requires reliable decomposition and temporal association from frozen features.
We therefore organize patch memory by temporal role to preserve complementary historical information within a compact memory.
The Structured Memory Bank converts the full context sequence into a compact set of memory tokens for subsequent retrieval.

Let
\begin{equation}
  U = \operatorname{TubeletView}(X_p,\mathcal{M}_c)
\end{equation}
denote the temporally ordered view of valid projected context tokens under context mask $\mathcal{M}_c$, with $U_{\tau}\in\mathbb{R}^{P\times D}$ containing the patch tokens at context tubelet $\tau$.
We divide the context into temporal index sets $\mathcal{I}_{\mathrm{anchor}}$, $\mathcal{I}_{\mathrm{middle}}$, and $\mathcal{I}_{\mathrm{recent}}$.
The anchor set covers the earliest context tubelets, the recent set covers the last visible tubelets before prediction, and the middle set covers intermediate history.
When $U$ provides a valid and temporally regular tubelet layout, the bank constructs four memory groups:
\begin{equation}
  M = [M_{\mathrm{anchor}}; M_{\mathrm{middle}};
  M_{\mathrm{recent}}; M_{\mathrm{global}}].
\end{equation}

Anchor memory stores the earliest context evidence, such as initial object identity, color, and position.
Middle memory compresses intermediate motion and interaction history.
Recent memory preserves the state immediately before prediction.
Global memory summarizes the full context at a coarse level.
Algorithm~\ref{alg:rrpm} specifies the construction of the four memory groups and their concatenation into $M$.
The operators $\rho_{\mathrm{anchor}}$, $\rho_{\mathrm{middle}}$, $\rho_{\mathrm{recent}}$, and $\rho_{\mathrm{global}}$ are lightweight token selection, projection, or compression operators that output a fixed number of memory tokens.
The learned role embeddings $e_{\mathrm{anchor}}$, $e_{\mathrm{middle}}$, $e_{\mathrm{recent}}$, and $e_{\mathrm{global}}$ identify the temporal function of each memory group and are broadcast across the corresponding tokens.
In our implementation, anchor and recent memory are selected from temporally designated tubelets, middle memory is compressed from grouped intermediate tubelets, and global memory is generated from the concatenated memory tokens.
All memory tokens live in the predictor hidden dimension, so they can be read without changing the pretrained encoder or target encoder.

If the context mask cannot be converted into a valid, temporally regular tubelet view, temporal partitioning is not reliable.
In this case, the bank uses a global-only fallback,
\begin{equation}
  M = \rho_{\mathrm{global}}(X_p) + e_{\mathrm{global}},
\end{equation}
where the global summarizer is applied directly to the projected valid context tokens.
This fallback preserves a well-defined memory interface without assigning potentially incorrect anchor, middle, or recent roles.

The temporal organization is asymmetric by design.
Anchor memory protects early evidence from being overwritten by later visual context.
Recent memory preserves the immediate pre-prediction state, which is useful for short-term motion continuity.
Middle memory gives the predictor access to interaction history without storing every patch token.
Global memory provides a coarse fallback when the exact temporal source of relevant evidence is ambiguous.
The bank stores patch-level evidence in a structured form without assuming that each memory token is a stable object.

\begin{figure*}[t]
  \centering
  \includegraphics[width=\textwidth]{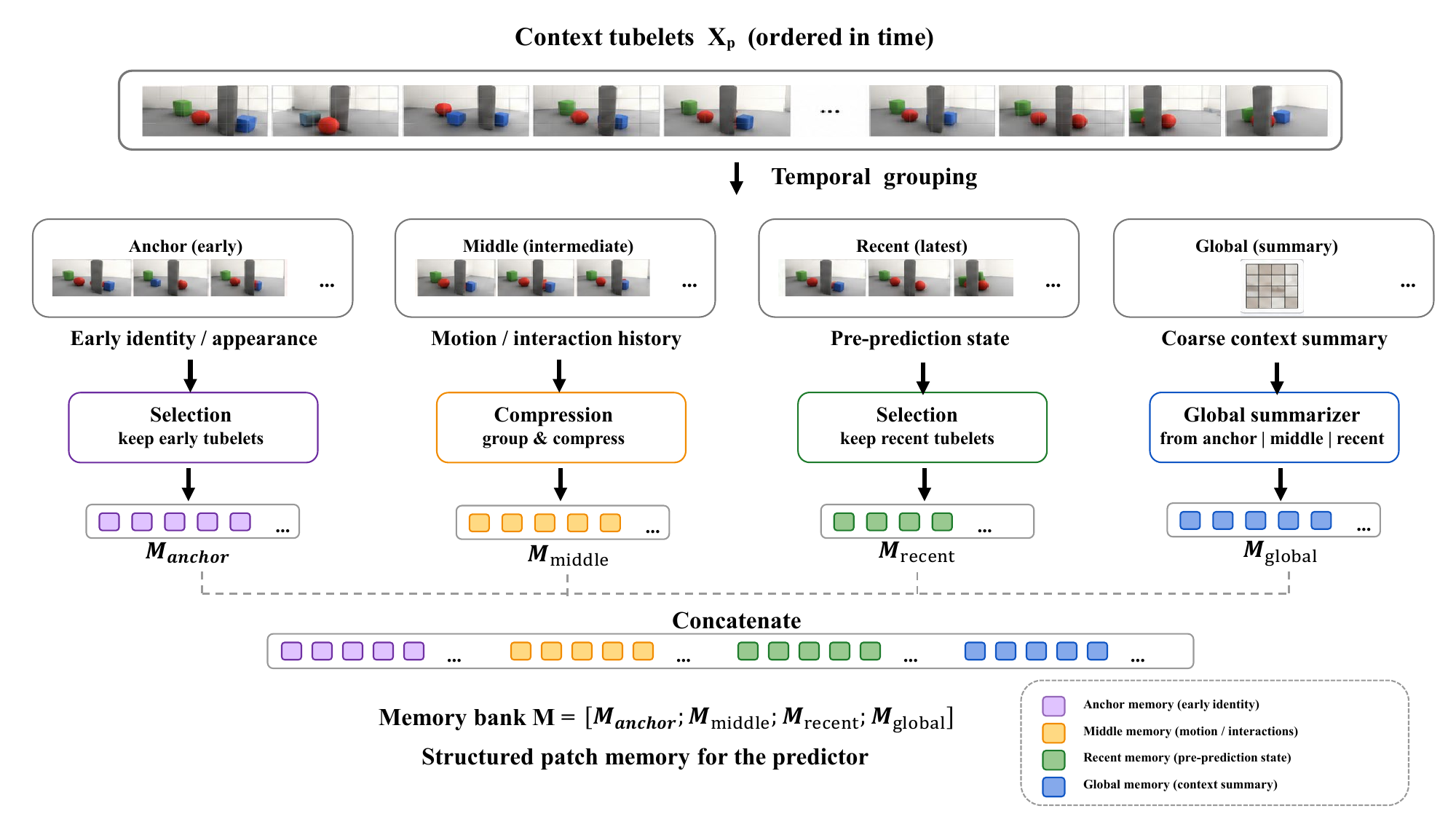}
  \caption{\textbf{Structured Memory Bank in RRPM.}
    The Structured Memory Bank preserves historical patch evidence according to temporal role.
    Anchor memory stores early object evidence, middle memory summarizes motion and interaction history, recent memory stores the state immediately before prediction, and global memory provides a coarse context summary.
  The bank gives the predictor access to physically relevant history without requiring object supervision.}
  \label{fig:patchbank}
\end{figure*}

\subsubsection{Memory Registers and Workspace Registers}
\label{subsec:registers}

The predictor needs a stable memory interface.
Directly modifying context tokens can disturb the pretrained prediction path.
A single shared register set can also mix two different roles: internal workspace and explicit memory carrier.
We split the register tokens into Workspace Registers $R_w$ and Memory Registers $R_{\mathrm{mem}}$.
Let $R_w\in\mathbb{R}^{N_w\times D}$ and $R_{\mathrm{mem}}\in\mathbb{R}^{N_{\mathrm{mem}}\times D}$, where $N_w$ and $N_{\mathrm{mem}}$ are the numbers of Workspace Registers and Memory Registers, respectively.
The predictor input is
\begin{equation}
  H_0 = [X_p; Y_{\mathrm{mask}}; R_w; R_{\mathrm{mem}}],
\end{equation}
where $Y_{\mathrm{mask}}$ denotes the future-mask tokens.
Workspace Registers participate only in predictor self-attention.
Memory Registers additionally read the Structured Memory Bank.

At read layers $\mathcal{L}_{\mathrm{read}}$, Memory Registers attend to $M$ through gated cross-attention:
\begin{equation}
  \begin{aligned}
    \Delta R_{\mathrm{mem}}^{(\ell)} ={}&
    \sigma(g_\ell)\operatorname{CrossAttn} (R_{\mathrm{mem}}^{(\ell)}, M, M), \\
    R_{\mathrm{mem}}^{(\ell)+} ={}&
    R_{\mathrm{mem}}^{(\ell)}+\Delta R_{\mathrm{mem}}^{(\ell)},
    \quad \ell\in\mathcal{L}_{\mathrm{read}}.
  \end{aligned}
\end{equation}
Only the Memory Registers are directly updated by this read operation.
The context tokens and future mask tokens receive memory information later through frozen predictor self-attention.
This delayed propagation lets memory enter through a small set of trainable carriers instead of rewriting the full context representation.

The gate $g_\ell$ is initialized conservatively so that training begins near the frozen predictor behavior.
In all reported RRPM experiments, $\mathcal{L}_{\mathrm{read}}=\{3,7\}$ under the zero-based block indexing used in Algorithm~\ref{alg:rrpm}; the reads therefore occur after the fourth and eighth predictor blocks.
This placement leaves several subsequent frozen predictor blocks in which retrieved evidence can be integrated.
The split-register design controls both where memory is read and which tokens carry it.
After the final predictor block, only the predicted future target tokens are
used to compute the latent prediction loss and surprise score, while the
register tokens are discarded.
Let $\operatorname{Norm}_{\mathrm{out}}$ and $W_{\mathrm{out}}$ denote the
frozen final normalization and output projection of the pretrained predictor,
respectively. Here, $\operatorname{TargetSlice}$ selects the positions
corresponding to future-mask tokens.
\begin{algorithm}[t]
  \caption{HERA with Register-Routed Patch Memory}
  \label{alg:rrpm}
  \begin{algorithmic}[1]
    \REQUIRE Inputs $X_p$, $Y_{\mathrm{mask}}$, and context mask $\mathcal{M}_c$
    \REQUIRE Frozen predictor blocks $B_{0:L-1}$
    \REQUIRE Trainable registers $R_w,R_{\mathrm{mem}}$, memory operators $\rho$, role embeddings $e$, and gates $g$
    \REQUIRE Temporal partitions $\mathcal{I}$ and zero-based read indices $\mathcal{L}_{\mathrm{read}}$
    \STATE $U \leftarrow \operatorname{TubeletView}(X_p,\mathcal{M}_c)$
    \IF{$U$ is invalid or temporally irregular}
    \STATE $M \leftarrow \rho_{\mathrm{global}}(X_p)+e_{\mathrm{global}}$
    \ELSE
    \STATE $M_{\mathrm{anchor}} \leftarrow \rho_{\mathrm{anchor}}(\{U_\tau:\tau\in\mathcal{I}_{\mathrm{anchor}}\})+e_{\mathrm{anchor}}$
    \STATE $M_{\mathrm{middle}} \leftarrow \rho_{\mathrm{middle}}(\{U_\tau:\tau\in\mathcal{I}_{\mathrm{middle}}\})+e_{\mathrm{middle}}$
    \STATE $M_{\mathrm{recent}} \leftarrow \rho_{\mathrm{recent}}(\{U_\tau:\tau\in\mathcal{I}_{\mathrm{recent}}\})+e_{\mathrm{recent}}$
    \STATE $M_{\mathrm{global}} \leftarrow \rho_{\mathrm{global}}([M_{\mathrm{anchor}};M_{\mathrm{middle}};M_{\mathrm{recent}}])+e_{\mathrm{global}}$
    \STATE $M \leftarrow [M_{\mathrm{anchor}};M_{\mathrm{middle}};M_{\mathrm{recent}};M_{\mathrm{global}}]$
    \ENDIF
    \STATE $H \leftarrow [X_p;Y_{\mathrm{mask}};R_w;R_{\mathrm{mem}}]$
    \FOR{$\ell=0$ \TO $L-1$}
    \STATE $H \leftarrow B_\ell(H)$
    \IF{$\ell\in\mathcal{L}_{\mathrm{read}}$}
    \STATE Extract $R_{\mathrm{mem}}$ from $H$
    \STATE $\Delta R_{\mathrm{mem}} \leftarrow \sigma(g_\ell)
    \operatorname{CrossAttn}(R_{\mathrm{mem}},M,M)$
    \STATE $R_{\mathrm{mem}} \leftarrow R_{\mathrm{mem}}+\Delta R_{\mathrm{mem}}$
    \STATE Replace the slice of $H$ corresponding to $R_{\mathrm{mem}}$
    \ENDIF
    \ENDFOR
    \STATE $\hat{Z}\leftarrow W_{\mathrm{out}}\bigl(\operatorname{TargetSlice}(\operatorname{Norm}_{\mathrm{out}}(H))\bigr)$
    \RETURN $\hat{Z}$
  \end{algorithmic}
\end{algorithm}
\subsubsection{Training and Inference}
\label{subsec:training}

Training follows the base model's future-latent prediction objective.
For each clip, the frozen context encoder produces context tokens, the Structured Memory Bank stores historical evidence, the memory-augmented predictor outputs future target latents, and the frozen target encoder provides normalized targets.
The optimization objective is
\begin{equation}
  \mathcal{L}_{\mathrm{pred}} =
  \frac{1}{|\mathcal{T}|D_z}
  \sum_{t\in\mathcal{T}}
  \left\|\hat{z}_{t} - \bar{z}_{t}\right\|_1 .
\end{equation}
Only the Structured Memory Bank, cross-attention readers and gates, Workspace Registers, and Memory Registers are updated.
The V-JEPA 2 encoders and main predictor blocks remain fixed.
Freezing the encoders and main predictor blocks limits the source of improvement.
If the entire predictor were fine-tuned, gains could come from general adaptation rather than from the memory interface.

At inference, the same procedure is applied independently to each video window.
The model constructs $M$, injects memory through $R_{\mathrm{mem}}$ at the read layers,
predicts future target latents, and averages the resulting window losses to
obtain the video-level AvgSurprise score.
The pairwise margin and accuracy are then computed exactly as in the physical violation probe.
No IntPhys2-specific classifier, threshold, or pairwise calibration is introduced.
The comparison tests whether the memory mechanism changes physical violation sensitivity, rather than whether a new scorer is better calibrated.

\section{Experiments}
\label{sec:experiments}

\subsection{Experimental Setup}
\label{subsec:setup}

\paragraph{Datasets and evaluation protocol.}
We train HERA with Register-Routed Patch Memory (RRPM) on the
14,000-video Physion training split~\citep{xiang2024physion} and evaluate it
on the IntPhys2 Main split~\citep{bordes2025intphys2}, which contains 1,012
videos organized into 506 matched possible--impossible pairs. No IntPhys2
videos, pair identities, physical-validity labels, or condition annotations
are used during training.

Following the AvgSurprise protocol defined in the Problem Formulation
subsection, all methods use the same predefined evaluation settings and
scoring pipeline. All reported results use context lengths selected from the
same predefined set under the same protocol.

Each evaluation uses 48-frame clips at $384\times384$ resolution, a temporal
sampling step of 10, a sliding-window stride of 2, a batch size of 1, and
bfloat16 inference. AvgSurprise averages future-latent prediction losses over
all valid sliding windows, and pairwise accuracy is computed using the
official IntPhys2 scorer. Zero-padded windows are excluded, and no ties occur
in the reported evaluations.

\paragraph{Implementation details.}
We use V-JEPA 2-G as the frozen backbone. Its context encoder, target encoder,
and predictor blocks remain fixed, leaving 3.00M trainable parameters in the
Structured Memory Bank, gated cross-attention readers, Memory
Registers, and Workspace Registers. RRPM contains 12 Memory Registers
and eight Workspace Registers. The former retrieve evidence from the
Structured Memory Bank after the fourth and eighth predictor blocks, while
the latter support its subsequent integration through predictor
self-attention. The Structured Memory Bank contains 96 anchor tokens,
28 middle tokens, 64 recent tokens, and eight global tokens, giving
196 memory tokens in total.

All methods are trained across multiple random seeds and use the same data
split, scoring pipeline, software environment, and hardware allocation.
Experiments are implemented in PyTorch and conducted on a four-GPU node
equipped with NVIDIA H100 GPUs. Direct-to-Target Memory
follows the same data and training schedule but tunes 5.96M parameters because
it additionally unfreezes the final two predictor blocks and predictor
normalization.

\subsection{Experimental Results}
\label{subsec:results}

\paragraph{Comparison with alternative memory mechanisms.}
Table~\ref{tab:main-results} compares HERA with alternative memory designs
covering clip-local storage, cross-window propagation, hierarchical temporal
selection, shared-register routing, and direct memory injection. Each row
reports pairwise accuracy under the same evaluation protocol.

\begin{table}[t]
  \centering
  \small
  \setlength{\tabcolsep}{5pt}
  \begin{tabular}{lc}
    \toprule
    Method & Accuracy (\%) \\
    \midrule
    V-JEPA 2-G & 52.57 \\
    \midrule
    Clip-Local Patch Memory & 52.77 \\
    Prefix Cross-Window Memory & 52.77 \\
    Hierarchical Patch Memory & 52.96 \\
    Shared-Register Memory & 52.96 \\
    Direct-to-Target Memory & 53.56 \\
    \midrule
    \textbf{HERA with RRPM} & \textbf{54.35} \\
    \midrule
    HERA + low-LR continuation & 53.95 \\
    RRPM w/o Memory Regs. & 53.56 \\
    RRPM w/o Workspace Regs. & 53.56 \\
    \bottomrule
  \end{tabular}
  \caption{Pairwise AvgSurprise accuracy (\%) on the IntPhys2 Main split
  under the same evaluation protocol. The highest accuracy is bold.}
  \label{tab:main-results}
\end{table}
HERA with RRPM achieves the
highest overall accuracy, improving V-JEPA 2-G from 52.57\% to 54.35\%, an
absolute gain of 1.78 percentage points. It also outperforms Direct-to-Target
Memory, the strongest competing routing design, by 0.79 points. Notably,
Direct-to-Target Memory tunes 5.96M parameters by adapting the final predictor
blocks, whereas HERA trains only 3.00M parameters and preserves the pretrained
prediction pathway.

Clip-local, prefix, and hierarchical patch memories provide smaller gains,
indicating that increasing the amount, temporal range, or organization of
stored history is insufficient without an effective retrieval and integration
pathway. Shared-Register Memory also remains below HERA, suggesting that using
a single token set for both memory access and predictor-side computation may
introduce interference between the two functions.

\paragraph{Register-role ablation.}
The register-removal variants are evaluated from the same checkpoint. Removing either register group
reduces the accuracy from 53.95\% to 53.56\%, corresponding to a
0.39-point decrease. Memory Registers provide the explicit route for
querying the Structured Memory Bank, while Workspace Registers integrate the
retrieved information through the frozen predictor's native self-attention
pathway. The consistent reduction from either removal supports the
complementary roles of the two register groups within the trained RRPM model.

\paragraph{Subgroup analysis.}
Table~\ref{tab:subgroup-results} reports an IntPhys2 subgroup analysis spanning
difficulty, camera, and physical-principle categories in which prediction
particularly depends on earlier appearance, trajectory, or interaction
evidence. These subgroups place stronger demands on recovering earlier
appearance, trajectory, or interaction evidence.

\begin{table}[t]
  \centering
  \small
  \setlength{\tabcolsep}{3.8pt}
  \begin{tabular}{lccc}
    \toprule
    Subgroup & V-JEPA 2-G & HERA & Gain \\
    \midrule
    Hard
    & 48.81 & \textbf{57.74} & +8.93\% \\
    Fixed camera
    & 48.56 & \textbf{56.73} & +8.17\% \\
    Continuity / Fixed
    & 46.15 & \textbf{57.69} & +11.54\% \\
    Immutability / Fixed
    & 46.15 & \textbf{63.46} & +17.31\% \\
    Solidity
    & 47.26 & \textbf{55.48} & +8.22\% \\
    \bottomrule
  \end{tabular}
  \caption{Pairwise AvgSurprise accuracy (\%) on IntPhys2 subgroups spanning
  difficulty, camera, and physical-principle categories under the same protocol.}
  \label{tab:subgroup-results}
\end{table}

HERA produces particularly strong improvements when physically relevant
evidence appears earlier in the video and must later be recovered for
prediction. Accuracy increases by 8.93 points on Hard examples and 8.17
points under fixed cameras. Larger gains are observed on fixed-camera
continuity and immutability, where HERA improves the backbone by 11.54 and
17.31 points, respectively. Solidity also improves by 8.22 points, a pattern
consistent with improved access to historical spatial and interaction evidence
through RRPM.

Overall, HERA provides the strongest result across five alternative memory
mechanisms. The register-removal diagnostics further support the central
design of HERA: temporally structured memory preserves historical evidence,
Memory Registers selectively retrieve it, and Workspace Registers
integrate the retrieved information without modifying the pretrained predictor
blocks.

\section{Discussion}
\label{sec:discussion}

Our results show that physical prediction benefits from routing retained
evidence into the predictor when needed. HERA improves performance across the
evaluated memory mechanisms, with clear gains on Hard, fixed-camera,
continuity, immutability, and solidity cases that depend on earlier evidence.
The architectural controls support separate Memory Registers for retrieval
and Workspace Registers for integration through the frozen predictor. HERA
requires no object supervision, and extending it to richer structured states
and broader physical settings is promising.

\section{Conclusion}
\label{sec:conclusion}

We introduced HERA, a historical-evidence routing framework, and instantiated
it with RRPM, a lightweight adapter that routes structured historical patch
evidence into a frozen latent video predictor through dedicated Memory
Registers and Workspace Registers. On the IntPhys2 Main split, under the same
evaluation protocol, HERA with RRPM improves the pairwise
AvgSurprise accuracy of V-JEPA 2-G from 52.57\% to 54.35\%.

Comparisons with five alternative memory mechanisms and register-role
ablations support the importance of selective retrieval and separate
predictor-side integration. These results show that making historical evidence
accessible at the appropriate stage of prediction is a practical way to
improve physical reasoning in pretrained latent world models.

\bibliography{aaai2027}



\end{document}